\documentclass[11pt,a4paper]{article}
\usepackage[hyperref]{acl2017}
\usepackage{times}
\usepackage{amssymb}
\usepackage{amsmath}
\usepackage{mathtools}
\usepackage{latexsym}
\usepackage{graphicx}
\usepackage[noend]{algpseudocode}
\usepackage{url}
\usepackage[lined,boxed,commentsnumbered]{algorithm2e}
\usepackage{setspace}
\usepackage{multirow}
\usepackage{comment}
\usepackage{hhline}
\usepackage{array}

\newcolumntype{C}[1]{>{\centering\let\newline\\\arraybackslash\hspace{-5pt}}m{#1}}

\aclfinalcopy 
\def\aclpaperid{***} 

\title{Joint Modeling of Content and Discourse Relations in Dialogues}

\author{Kechen Qin$^{1}$ ~~~~ Lu Wang$^{1}$  ~~~~ Joseph Kim$^{2}$\\
  $^{1}$College of Computer and Information Science, Northeastern University\\
  $^{2}$Computer Science and Artificial Intelligence Laboratory, \\ Massachusetts Institute of Technology\\
{\tt $^{1}$qin.ke@husky.neu.edu},~{\tt luwang@ccs.neu.edu}\\
{\tt $^{2}$joseph\_kim@csail.mit.edu}
   \\}

\begin{document}
\maketitle

\begin{abstract}
\fontsize{10}{12}\selectfont
We present a joint modeling approach to identify salient discussion points in spoken meetings as well as to label the discourse relations between speaker turns. A variation of our model is also discussed when discourse relations are treated as latent variables. Experimental results on two popular meeting corpora show that our joint model can outperform state-of-the-art approaches for both phrase-based content selection and discourse relation prediction tasks. We also evaluate our model on predicting the consistency among team members' understanding of their group decisions. Classifiers trained with features constructed from our model achieve significant better predictive performance than the state-of-the-art.

\end{abstract}

\section{Introduction}
Goal-oriented dialogues, such as meetings, negotiations, or customer service transcripts, play an important role in our daily life. Automatically extracting the critical points and important outcomes from dialogues would facilitate generating summaries for complicated conversations, understanding the decision-making process of meetings, or analyzing the effectiveness of collaborations.

\begin{figure}[t]
\centering
\includegraphics[width=78mm,height=46mm]{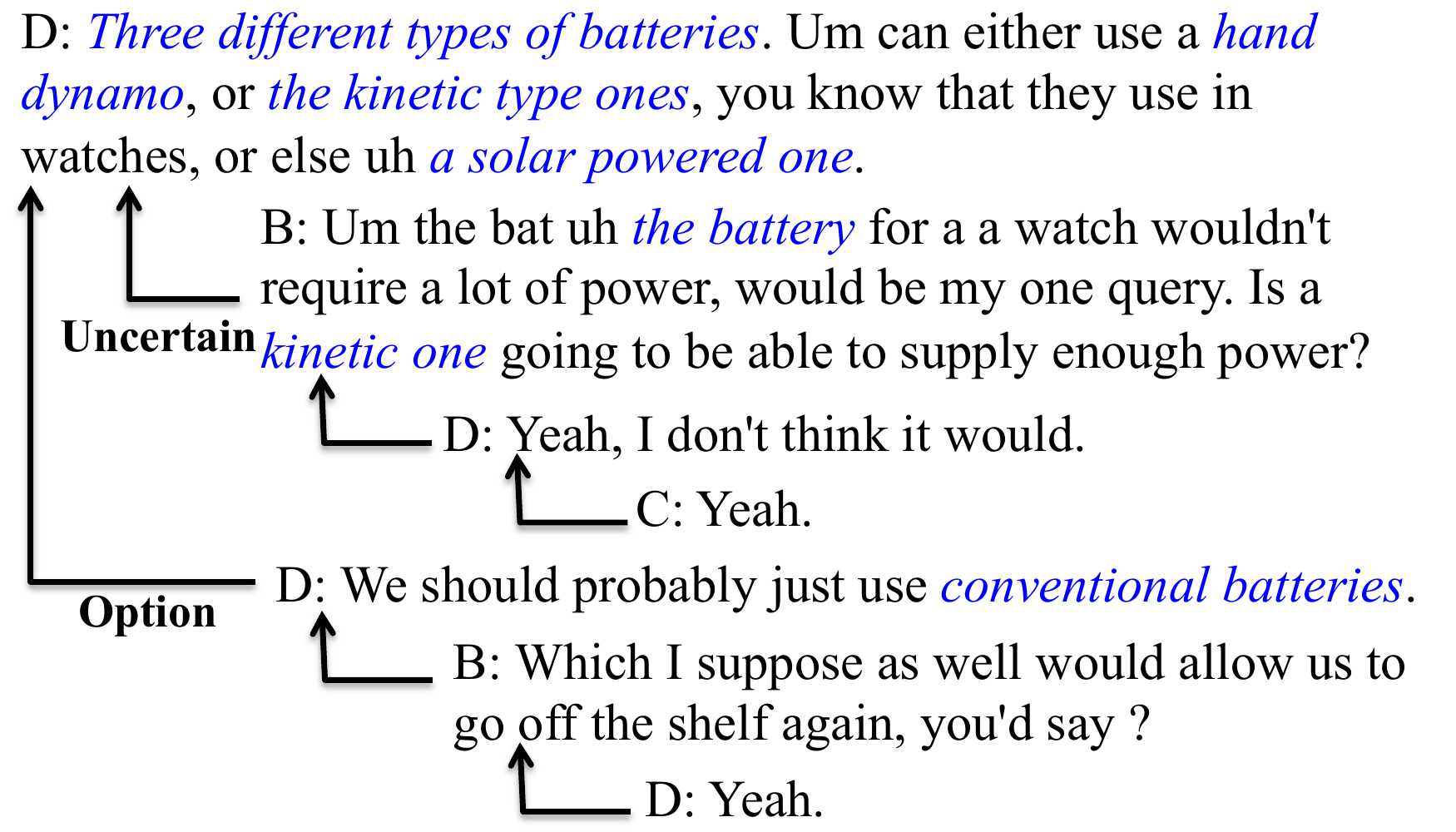}
\caption{\fontsize{10}{12}\selectfont A sample clip from AMI meeting corpus. B, C, and D denotes different speakers. Here we highlight salient phrases (in \textit{italics}) that are relevant to the major topic discussed, i.e., ``which type of battery to use for the remote control". Arrows indicate discourse structure between speaker turns. We also show some of the discourse relations for illustration.}
\label{fig:example_intro}
\end{figure}

We are interested in a specific type of dialogues --- spoken meetings, which is a common way for collaboration and idea sharing. Previous work~\cite{kirschner2012visualizing} has shown that discourse structure can be used to capture the main discussion points and arguments put forward during problem-solving and decision-making processes in meetings. Indeed, content of different speaker turns do not occur in isolation, and should be interpreted within the context of discourse. Meanwhile, content can also reflect the purpose of speaker turns, thus facilitate with discourse relation understanding. 
Take the meeting snippet from AMI corpus~\cite{ami} in Figure~\ref{fig:example_intro} as an example. This discussion is annotated with discourse structure based on the Twente Argumentation Schema (TAS) by~\newcite{so65562}, which focuses on argumentative discourse information. As can be seen, meeting participants evaluate different options by showing doubt (\textsc{uncertain}), bringing up alternative solution (\textsc{option}), or giving feedback. The discourse information helps with the identification of the key discussion point, i.e., ``which type of battery to use", by revealing the discussion flow.

To date, most efforts to leverage discourse information to detect salient content from dialogues have focused on encoding gold-standard discourse relations as features for use in classifier training~\cite{murray2006incorporating,Galley:2006:SCR:1610075.1610126,mckeown2007using,Bui:2009:EDM:1708376.1708410}. However, automatic discourse parsing in dialogues is still a challenging problem~\cite{perret-EtAl:2016:N16-1}. Moreover, acquiring human annotation on discourse relations is a time-consuming and expensive process, and does not scale for large datasets. 

In this paper, we propose \textit{a joint modeling approach to select salient phrases reflecting key discussion points as well as label the discourse relations between speaker turns in spoken meetings}. 
We hypothesize that leveraging the interaction between content and discourse has the potential to yield better prediction performance on both \textit{phrase-based content selection} and \textit{discourse relation prediction}. 
Specifically, we utilize argumentative discourse relations as defined in Twente Argument Schema (TAS)~\cite{so65562}, where discussions are organized into tree structures with discourse relations labeled between nodes (as shown in Figure~\ref{fig:example_intro}). 
Algorithms for joint learning and joint inference are proposed for our model. We also present a variation of our model to treat discourse relations as latent variables when true labels are not available for learning. 
We envision that the extracted salient phrases by our model can be used as input to abstractive meeting summarization systems~\cite{wang-cardie:2013:ACL2013,mehdad-carenini-ng:2014:P14-1}. Combined with the predicted discourse structure, a visualization tool can be exploited to display conversation flow to support intelligent meeting assistant systems.

To the best of our knowledge, our work is the first to jointly model content and discourse relations in meetings. We test our model with two meeting corpora --- the AMI corpus~\cite{ami} and the ICSI corpus~\cite{icsi}. 
Experimental results show that our model yields an accuracy of 63.2 on phrase selection, which is significantly better than a classifier based on Support Vector Machines (SVM). Our discourse prediction component also obtains better accuracy than a state-of-the-art neural network-based approach (59.2 vs. 54.2). 
%
Moreover, our model trained with latent discourse outperforms SVMs on both AMI and ICSI corpora for phrase selection. 
We further evaluate the usage of selected phrases as extractive meeting summaries. Results evaluated by ROUGE~\cite{Lin:2003:AES:1073445.1073465} demonstrate that our system summaries obtain a ROUGE-SU4 F1 score of 21.3 on AMI corpus, which outperforms non-trivial extractive summarization baselines and a keyword selection algorithm proposed in~\newcite{liu2009unsupervised}.

Moreover, since both content and discourse structure are critical for building shared understanding among participants~\cite{mulder2002assessing,mercer2004sociocultural}, we further investigate whether our learned model can be utilized to predict the consistency among team members' understanding of their group decisions. This task is first defined as \textit{consistency of understanding} (COU) prediction by~\newcite{kim2016improving}, who have labeled a portion of AMI discussions with consistency or inconsistency labels. 
We construct features from our model predictions to capture different discourse patterns and word entrainment scores for discussion with different COU level. 
Results on AMI discussions show that SVM classifiers trained with our features significantly outperform the state-of-the-art results~\cite{kim2016improving} (F1: 63.1 vs. 50.5) and non-trivial baselines.

The rest of the paper is structured as follows: we first summarize related work in Section~\ref{sec:related}. The joint model is presented in Section~\ref{sec:model}. Datasets and experimental setup are described in Section~\ref{sec:data}, which is followed by experimental results (Section~\ref{sec:result}). We then study the usage of our model for predicting consistency of understanding in groups in Section~\ref{sec:consistency}. We finally conclude in Section~\ref{sec:conclusion}.


\section{Related Work}
\label{sec:related}
Our model is inspired by research work that leverages discourse structure for identifying salient content in conversations, which is still largely reliant on features derived from gold-standard discourse labels~\cite{mckeown2007using,Murray:2010:GVA:1873738.1873753,bokaei2016extractive}. 
For instance, adjacency pairs, which are paired utterances with question-answer or offer-accept relations, are found to frequently appear in meeting summaries together and thus are utilized to extract summary-worthy utterances by~\newcite{Galley:2006:SCR:1610075.1610126}. 
%
There is much less work that jointly predicts the importance of content along with the discourse structure in dialogus. \newcite{oya2014extractive} employs Dynamic Conditional Random Field to recognize sentences in email threads for use in summary as well as their dialogue acts. Only local discourse structures from adjacent utterances are considered. Our model is built on tree structures, which captures more global information. 
Our work is also in line with keyphrase identification or phrase-based summarization for conversations. Due to the noisy nature of dialogues, recent work focuses on identifying summary-worthy phrases from meetings~\cite{FernandezFDAEP08,Riedhammer:2010:LSS:1837521.1837625} or email threads~\cite{loza2014building}. For instance, \newcite{wang-cardie:2012:SIGDIAL20122} treat the problem as an information extraction task, where summary-worthy content represented as indicator and argument pairs is identified by an unsupervised latent variable model. 
Our work also targets at detecting salient phrases from meetings, but focuses on the joint modeling of critical discussion points and discourse relations held between them.

For the area of discourse analysis in dialogues, a significant amount of work has been done in predicting local discourse structures, such as recognizing dialogue acts or social acts of adjacent utterances from phone conversations~\cite{stolcke2000dialogue,kalchbrenner2013recurrent,ji-haffari-eisenstein:2016:N16-1}, spoken meetings~\cite{dielmann2008recognition}, or emails~\cite{Cohen04learningto}. 
%
Although discourse information from non-adjacent turns has been studied in the context of online discussion forums~\cite{ghosh2014analyzing} and meetings~\cite{hakkani2009towards}, none of them models the effect of discourse structure on content selection, which is a gap that this work fills in.

\section{The Joint Model of Content and Discourse Relations}
\label{sec:model}
In this section, we first present our joint model in Section~\ref{sec:modeldescription}. The algorithms for learning and inference are described in Sections~\ref{sec:learning} and~\ref{sec:inference}, followed by feature description (Section~\ref{sec:feature}).

\subsection{Model Description}
\label{sec:modeldescription}
Our proposed model learns to jointly perform phrase-based content selection and discourse relation prediction by making use of the interaction between the two sources of information. 
Assume that a meeting discussion is denoted as $\mathbf{x}$, where $\mathbf{x}$ consists of a sequence of discourse units $\mathbf{x}=\{x_{1}, x_{2}, \cdots ,x_{n}\}$. Each discourse unit can be a complete speaker turn or a part of it. As demonstrated in Figure~\ref{fig:example_intro}, a tree-structured discourse diagram is constructed for each discussion with each discourse unit $x_{i}$ as a node of the tree. In this work, we consider the argumentative discourse structure by Twente Argument Schema (TAS)~\cite{so65562}. 
For each node $x_{i}$, it is attached to another node $x_{i^\prime}$ ($i^\prime<i$) in the discussion, and a discourse relation $d_{i}$ is hold on the link $\langle x_{i}, x_{i^\prime} \rangle$ ($d_{i}$ is empty if $x_{i}$ is the root). Let $\mathbf{t}$ denote the set of links $\langle x_{i}, x_{i^\prime} \rangle$ in $\mathbf{x}$.
Following previous work on discourse analysis in meetings~\cite{so65562,hakkani2009towards}, we assume that the attachment structure between discourse units are given during both training and testing.

A set of candidate phrases are extracted from each discourse unit $x_{i}$, from which salient phrases that contain gist information will be identified. We obtain constituent and dependency parses for utterances using Stanford parser~\cite{Klein:2003:AUP:1075096.1075150}. We restrict eligible candidate to be a noun phrase (NP), verb phrase (VP), prepositional phrase (PP), or adjective phrase (ADJP) with at most 5 words, and its head word cannot be a stop word.\footnote{Other methods for mining candidate phrases, such as frequency-based method~\cite{liu2015mining}, will be studied for future work.} 
If a candidate is a parent of another candidate in the constituent parse tree, we will only keep the parent. We further merge a verb and a candidate noun phrase into one candidate if the later is the direct object or subject of the verb. For example, from utterance ``let's use a rubber case as well as rubber buttons", we can identify candidates ``use a rubber case" and ``rubber buttons". 
For $x_{i}$, the set of candidate phrases are denoted as $c_{i}=\{c_{i,1},c_{i,2},\cdots,c_{i,m_i}\}$, where $m_i$ is the number of candidates. $c_{i,j}$ takes a value of $1$ if the corresponding candidate is selected as salient phrase; otherwise, $c_{i,j}$ is equal to $0$. All candidate phrases in discussion $\mathbf{x}$ are represented as $\mathbf{c}$.


We then define a log-linear model with feature parameters $\mathbf{w}$ for the candidate phrases $\mathbf{c}$ and discourse relations $\mathbf{d}$ in $\mathbf{x}$ as:

{\fontsize{10}{10}\selectfont
\begin{equation}
\begin{split}
& p(\mathbf{c}, \mathbf{d}|\mathbf{x}, \mathbf{w}) \propto \exp [\mathbf{w}\cdot \Phi (\mathbf{c}, \mathbf{d}, \mathbf{x})]\\
& \propto \exp [\mathbf{w}\cdot \sum_{i=1, <x_i, x_{i^\prime}>\in \mathbf{t}}^{n} \phi (c_{i}, d_{i}, d_{i^\prime}, \mathbf{x})]\\
& \propto \exp [\sum_{i=1,<x_i, x_{i^\prime}>\in \mathbf{t}}^{n} ( \mathbf{w_c}\cdot \sum_{j=1}^{m_i} \phi_c (c_{i,j}, \mathbf{x})\\ 
&+ \mathbf{w_d}\cdot \phi_d (d_{i},d_{i^\prime}, \mathbf{x}) + \mathbf{w_{cd}}\cdot \sum_{j=1}^{m_i} \phi_{cd} (c_{i,j}, d_{i}, \mathbf{x}) ) ]\\
\end{split}
\end{equation}
\label{eq:objective}
}
Here $\Phi(\cdot)$ and $\phi(\cdot)$ denote feature vectors. We utilize three types of feature functions: (1) content-only features $\phi_c(\cdot)$, which capture the importance of phrases, (2) discourse-only features $\phi_d(\cdot)$, which characterize the (potentially higher-order) discourse relations, and (3) joint features of content and discourse $\phi_{cd}(\cdot)$, which model the interaction between the two. $\mathbf{w_c}$, $\mathbf{w_d}$, and $\mathbf{w_{cd}}$ are corresponding feature parameters. Detailed feature descriptions can be found in Section~\ref{sec:feature}. 

\noindent \textbf{Discourse Relations as Latent Variables.} 
As we mentioned in the introduction, acquiring labeled training data for discourse relations is a time-consuming process since it would require human annotators to inspect the full discussions. Therefore, we further propose a variation of our model where it treats the discourse relations as latent variables, so that $p(\mathbf{c} |\mathbf{x}, \mathbf{w})=\sum_{\mathbf{d}} p(\mathbf{c}, \mathbf{d}|\mathbf{x}, \mathbf{w})$. 
Its learning algorithm is slightly different as described in the next section. 


\subsection{Joint Learning for Parameter Estimation}
\label{sec:learning}
For learning the model parameters $\mathbf{w}$, we employ an algorithm based on SampleRank~\cite{rohanimanesh2011samplerank}, which is a stochastic structure learning method. In general, the learning algorithm constructs a sequence of configurations for sample labels as a Markov chain Monte Carlo (MCMC) chain based on a task-specific loss function, where stochastic gradients are distributed across the chain. 

The full learning procedure is described in Algorithm~\ref{alg:learning}. To start with, the feature weights $\mathbf{w}$ is initialized with each value randomly drawn from $[-1, 1]$. Multiple epochs are run through all samples. 
For each sample, we randomly initialize the assignment of candidate phrases labels $\mathbf{c}$ and discourse relations $\mathbf{d}$. 
Then an MCMC chain is constructed with a series of configurations $\sigma =(\mathbf{c}$, $\mathbf{d})$: at each step, it first samples a discourse structure $\mathbf{d}$ based on the proposal distribution $q(\mathbf{d^\prime} |\mathbf{d},\mathbf{x})$, and then samples phrase labels conditional on the new discourse relations and previous phrase labels based on $q(\mathbf{c^\prime} |\mathbf{c}, \mathbf{d^\prime},\mathbf{x})$. Local search is used for both proposal distributions.\footnote{For future work, we can explore other proposal distributions that utilize the conditional distribution of salient phrases given sampled discourse relations.}   
The new configuration is accepted if it improves on the score by $\omega (\sigma ^\prime)$. The parameters $\mathbf{w}$ are updated accordingly.

For the scorer $\omega$, we use a weighted combination of F1 scores of phrase selection ($F1_{c}$) and discourse relation prediction ($F1_{d}$): $\omega (\sigma)= \alpha \cdot F1_{c} +  (1-\alpha) \cdot F1_{d}$. We fix $\alpha$ to $0.1$.

When discourse relations are treated as latent, we initialize discourse relations for each sample with a label in $\{1, 2, \ldots, K\}$ if there are $K$ relations indicated, and we only use $F1_{c}$ as the scorer.

\begin{algorithm}[t]
{
\setstretch{0.3}
\fontsize{9}{10}\selectfont
\SetKwInOut{Input}{Input}
\SetKwInOut{Output}{Output}

\Input{$\mathbf{X}=\{\mathbf{x}\}$: discussions in the training set, \\
	$\eta$: learning rate, 
	$\epsilon$: number of epochs, \\
	$\delta$: number of sampling rounds, \\
	$\omega (\cdot)$: scoring function,
	$\Phi (\cdot)$: feature functions
	}

\Output{feature weights $\frac{1}{|\mathcal{W}|}\sum_{\mathbf{w}\in \mathcal{W}} \mathbf{w}$}
\BlankLine

Initialize $\mathbf{w}$\;
$\mathcal{W} \leftarrow \{\mathbf{w} \}$\;

\For {$e=1$ to $\epsilon$} {

	\For {$\mathbf{x}$ in $\mathbf{X}$} {
	
		\tcp{Initialize configuration for $\mathbf{x}$}
		Initialize $\mathbf{c}$ and $\mathbf{d}$\;
		$\sigma=(\mathbf{c}, \mathbf{d})$\;
		\For {$s=1$ to $\delta$} {
			\tcp{New configuration via local search}
			$\mathbf{d^\prime} \sim q_d(\cdot | \mathbf{x}, \mathbf{d})$\;
			$\mathbf{c^\prime} \sim q_d(\cdot | \mathbf{x}, \mathbf{c}, \mathbf{d^\prime})$\;
			
			$\sigma^\prime=(\mathbf{c^\prime}, \mathbf{d^\prime})$\;
			
			$\sigma^+=\arg \max_{\tilde{\sigma} \in \{\sigma, \sigma^\prime \}} \omega (\tilde{\sigma}) $\;
			$\sigma^-=\arg \min_{\tilde{\sigma} \in \{\sigma, \sigma^\prime \}} \omega (\tilde{\sigma}) $\;		
			
			$\hat{\nabla}=\Phi (\sigma^+)- \Phi (\sigma^-)$\;
			$\Delta \omega= \omega (\sigma^+)- \omega (\sigma^-)$\; 

			\tcp{Update parameters}			
			\If {$\mathbf{w}\cdot \hat{\nabla} < \Delta \omega$ \& $\Delta \omega\neq 0$}{
				$\mathbf{w} \leftarrow \mathbf{w}+\eta \cdot \hat{\nabla}$\;
				Add $\mathbf{w}$ in $\mathcal{W}$\;
			}
			
			\tcp{Accept or reject new configuration}
			\If {$\sigma^+ == \sigma^\prime$}{
				$\sigma=\sigma^\prime$
			
			}
		}
	
	}
}
}

\caption{\fontsize{10}{12}\selectfont  SampleRank-based joint learning.}
\label{alg:learning}
\end{algorithm}

\subsection{Joint Inference for Prediction}
\label{sec:inference}
Given a new sample $\mathbf{x}$ and learned parameters $\mathbf{w}$, we predict phrase labels and discourse relations as $\arg \max_{\mathbf{c}, \mathbf{d}} p(\mathbf{c}, \mathbf{d}|\mathbf{x}, \mathbf{w})$.

%
Dynamic programming can be employed to carry out joint inference, however, it would be time-consuming since our objective function has a large search space for both content and discourse labels. Hence we propose an alternating optimizing algorithm to search for $\mathbf{c}$ and $\mathbf{d}$ iteratively. Concretely, for each iteration, we first optimize on $\mathbf{d}$ by maximizing $\sum_{i=1,<x_i, x_i^\prime>\in \mathbf{t}}^{n} (\mathbf{w_d}\cdot \phi_d (d_{i},d_{i^\prime}, \mathbf{x}) + \mathbf{w_{cd}}\cdot \sum_{j=1}^{m_i} \phi_{cd} (c_{i,j}, d_{i}, \mathbf{x}))$. Message-passing~\cite{smith2008dependency} is used to find the best $\mathbf{d}$.

In the second step, we search for $\mathbf{c}$ that maximizes $\sum_{i=1,<x_i, x_i^\prime>\in \mathbf{t}}^{n} (\mathbf{w_c}\cdot \sum_{j=1}^{m_i} \phi_c (c_{i,j}, \mathbf{x}) + \mathbf{w_{cd}}\cdot \sum_{j=1}^{m_i} \phi_{cd} (c_{i,j}, d_{i}, \mathbf{x}) )$. We believe that candidate phrases based on the same concepts should have the same predicted label. Therefore, candidates of the same phrase type and sharing the same head word are grouped into one cluster. We then cast our task as an integer linear programming problem.\footnote{We use lpsolve: \url{http://lpsolve.sourceforge.net/5.5/}.} We optimize our objective function under constraints: (1) $c_{i,j}=c_{i^\prime, j^\prime}$ if $c_{i,j}$ and $c_{i^\prime, j^\prime}$ are in the same cluster, and (2) $c_{i,j}\in \{0, 1\} $, $\forall i, j$. 

The inference process is the same for models trained with latent discourse relations.

\subsection{Features}
\label{sec:feature}
We use features that characterize content, discourse relations, and the combination of both.

\noindent \textbf{Content Features.} 
For modeling the salience of content, we calculate the minimum, maximum, and average of \texttt{TF-IDF} scores of words and \texttt{number of content words} in each phrase based on the intuition that important phrases tend to have more content words with high TF-IDF scores~\cite{FernandezFDAEP08}. We also consider whether the head word of the phrase has been \texttt{mentioned in preceding turn}, which implies the focus of a discussion. The \texttt{size of the cluster} each phrase belongs to is also included. 
\texttt{Number of POS tags} and \texttt{phrase types} are counted to characterize the syntactic structure. Previous work~\cite{wang-cardie:2012:SIGDIAL20122} has found that a discussion usually ends with decision-relevant information. We thus identify the \texttt{absolute and relative positions} of the turn containing the candidate phrase in the discussion. Finally, we record whether the candidate phrase is \texttt{uttered by the main speaker}, who speakers the most words in the discussion.

\noindent \textbf{Discourse Features.} 
For each discourse unit, we collect the \texttt{dialogue act types} of the current unit and its parent node in discourse tree, whether there is any \texttt{adjacency pair} held between the two nodes~\cite{hakkani2009towards}, and the \texttt{Jaccard similarity} between them.
We record whether two turns are \texttt{uttered by the same speaker}, for example, \textsc{elaboration} is commonly observed between the turns from the same participant. 
We also calculate the \texttt{number of candidate phrases} based on the observation that \textsc{option} and \textsc{specialization} tend to contain more informative words than \textsc{positive} feedback. 
Length of the discourse unit is also relevant. Therefore, we compute the \texttt{time span} and \texttt{number of words}.  
To incorporate global structure features, we encode the \texttt{depth of the node} in the discourse tree and \texttt{the number of its siblings}. 
Finally, we include an \texttt{order-2 discourse relation} feature that encodes the relation between current discourse unit and its parent, and the relation between the parent and its grandparent if it exists.

\noindent \textbf{Joint Features.} For modeling the interaction between content and discourse, the discourse relation is added to each content feature to compose a joint feature. For example, if candidate $c$ in discussion $x$ has a content feature $\phi_{[avg-TFIDF]} (c, \mathbf{x})$ with a value of $0.5$, and its discourse relation $d$ is \textsc{positive}, then the joint feature takes the form of $\phi_{[avg-TFIDF, Positive]} (c, d, \mathbf{x})=0.5$.

\section{Datasets and Experimental Setup}
\label{sec:data}
\noindent \textbf{Meeting Corpora.} 
We evaluate our joint model on two meeting corpora with rich annotations: the AMI meeting corpus~\cite{ami} and the ICSI meeting corpus~\cite{icsi}. 
AMI corpus consists of 139 scenario-driven meetings, and ICSI corpus contains 75 naturally occurring meetings. 
%
Both of the corpora are annotated with dialogue acts, adjacency pairs, and topic segmentation. We treat each topic segment as one discussion, and remove discussions with less than 10 turns or labeled as ``opening" and ``chitchat". 
694 discussions from AMI and 1139 discussions from ICSI are extracted, and these two datasets are henceforth referred as \textsc{AMI-full} and \textsc{ICSI-full}.

\noindent \textbf{Acquiring Gold-Standard Labels.} 
Both corpora contain human constructed abstractive summaries and extractive summaries on meeting level. Short abstracts, usually in one sentence, are constructed by meeting participants --- \textit{participant summaries}, and external annotators --- \textit{abstractive summaries}. Dialogue acts that contribute to important output of the meeting, e.g. decisions, are identified and used as extractive summaries, and some of them are also linked to the corresponding abstracts. 

Since the corpora do not contain phrase-level importance annotation, we induce gold-standard labels for candidate phrases based on the following rule. A candidate phrase is considered as a positive sample if its head word is contained in any abstractive summary or participant summary. On average, 71.9 candidate phrases are identified per discussion for \textsc{AMI-full} with 31.3\% labeled as positive, and 73.4 for \textsc{ICSI-full} with 24.0\% of them as positive samples.

Furthermore, a subset of discussions in \textsc{AMI-full} are annotated with discourse structure and relations based on Twente Argumentation Schema (TAS) by~\newcite{so65562}\footnote{There are 9 types of relations in TAS: \textsc{positive}, \textsc{negative}, \textsc{uncertain}, \textsc{request}, \textsc{specialization}, \textsc{elaboration}, \textsc{option}, \textsc{option exclusion}, and \textsc{subject-to}.}. 
A tree-structured argument diagram (as shown in Figure~\ref{fig:example_intro}) is created for each discussion or a part of the discussion. The nodes of the tree contain partial or complete speaker turns, and discourse relation types are labeled on the links between the nodes. 
In total, we have 129 discussions annotated with discourse labels. 
This dataset is called \textsc{AMI-sub} hereafter. 



\noindent \textbf{Experimental Setup.} 
5-fold cross validation is used for all experiments. All real-valued features are uniformly normalized to [0,1]. 
For the joint learning algorithm, we use 10 epochs and carry out 50 sampling for MCMC for each training sample. The learning rate is set to 0.01. We run the learning algorithm for 20 times, and use the average of the learned weights as the final parameter values. 
For models trained with latent discourse relations, we fix the number of relations to $9$.

\noindent \textbf{Baselines and Comparisons.} 
For both phrase-based content selection and discourse relation prediction tasks, we consider a baseline that always predicts the majority label (Majority). 
Previous work has shown that Support Vector Machines (SVMs)-based classifiers achieve state-of-the-art performance for keyphrase selection in meetings~\cite{FernandezFDAEP08,wang-cardie:2013:ACL2013} and discourse parsing for formal text~\cite{HernaultPdI10}. Therefore, we compare with linear SVM-based classifiers, trained with the same feature set of content features or discourse features. We fix the trade-off parameter to $1.0$ for all SVM-based experiments. For discourse relation prediction, we use one-vs-rest strategy to build multiple binary classifiers.\footnote{Multi-class classifier was also experimented with, but gave inferior performance.} 
We also compare with a state-of-the-art discourse parser~\cite{ji-haffari-eisenstein:2016:N16-1}, which employs neural language model to predict discourse relations.

%

\section{Experimental Results}
\label{sec:result}
\subsection{Phrase Selection and Discourse Labeling}

Here we present the experimental results on phrase-based content selection and discourse relation prediction. 
We experiment with two variations of our joint model: one is trained on gold-standard discourse relations, the other is trained by treating discourse relations as latent models as described in Section~\ref{sec:modeldescription}. Remember that we have gold-standard argument diagrams on the \textsc{AMI-sub} dataset, we can thus conduct experiments by assuming the \textit{True Attachment Structure} is given for latent versions. When argument diagrams are not available, we build a tree among the turns in each discussion as follows. Two turns are attached if there is any adjacency pair between them. If one turn is attached to more than one previous turns, the closest one is considered. For the rest of the turns, they are attached to the preceding turn. This construction is applied on \textsc{AMI-full} and \textsc{ICSI-full}. 

\begin{table}[t]
{
	\centering
	\fontsize{9}{10}\selectfont
    \setlength{\tabcolsep}{3.0mm}
	
	\begin{tabular}{|l|l|l|}
 	\hline
  		& \textbf{Acc} & \textbf{F1} \\
 	\hline 
 	\underline{\textbf{Comparisons}} & & \\
 
 	Baseline (Majority) & 60.1 & 37.5\\ 
 	SVM (w content features in \S~\ref{sec:feature}) &  57.8 & 54.6 \\ 
	\hline
	\hline 	
 	
 	\underline{\textbf{Our Models}} & & \\
 	Joint-Learn + Joint-Inference & {\bf 63.2}$\ast$ & {\bf 62.6}$\ast$\\ 
 	Joint-Learn + Separate-Inference & 57.9 & 57.8\\
 	Separate-Learn & 53.4 & 52.6\\
 	
	\hline
	\hline 	
 	
	\underline{\textbf{Our Models (Latent Discourse)}} & & \\
	\textit{w/ True Attachment Structure} & & \\
	Joint-Learn + Joint-Inference & 60.3$\ast$ & 60.3$\ast$ \\ 
	Joint-Learn + Separate-Inference & 56.4 & 56.2 \\

	\textit{w/o True Attachment Structure} & & \\
	Joint-Learn + Joint-Inference & 56.4 & 56.4 \\ 
	Joint-Learn + Separate-Inference & 52.7 & 52.3 \\ 

 \hline
\end{tabular}
}
\caption{\fontsize{10}{12}\selectfont  Phrase-based content selection performance on \textsc{AMI-sub} with accuracy (acc) and F1. We display results of our models trained with gold-standard discourse relation labels and with latent discourse relations. For the later, we also show results based on \textit{True Attachment Structure}, where the gold-standard attachments are known, and without the \textit{True Attachment Structure}. Our models that significantly outperform SVM-based model are highlighted with $\ast$ ($p < 0.05$, paired $t$-test). Best result for each column is in \textbf{bold}.}
\label{tab:ami_disc_phrase}
\end{table}

\begin{table}[t]
{
	\centering
	\fontsize{9}{10}\selectfont
    \setlength{\tabcolsep}{2.2mm}
	\begin{tabular}{|l|l|l|}
 	\hline
  		& \textbf{Acc} & \textbf{F1} \\
 	\hline
 	\underline{\textbf{Comparisons}} & & \\
 
 	Baseline (Majority) & 51.2 & 7.5\\ 
 	SVM (w discourse features in \S~\ref{sec:feature}) &  51.2 & 22.8 \\ 
 	\newcite{ji-haffari-eisenstein:2016:N16-1} & 54.2 & 21.4 \\
	\hline
	\hline 	
 	
 	\underline{\textbf{Our Models}} & & \\
 	Joint-Learn + Joint-Inference & 58.0$\ast$ & 21.7\\ 
 	Joint-Learn + Separate-Inference & {\bf 59.2}$\ast$ & 23.4\\
 	Separate-Learn & 58.2$\ast$ & {\bf 25.1}\\
	
 \hline
\end{tabular}

}
\caption{\fontsize{10}{12}\selectfont  Discourse relation prediction performance on \textsc{AMI-sub}. Our models that significantly outperform SVM-based model and \newcite{ji-haffari-eisenstein:2016:N16-1} are highlighted with $\ast$ ($p < 0.05$, paired $t$-test). Best result for each column is in \textbf{bold}.
}
\label{tab:ami_disc_discourse}
\end{table}

We also investigate whether joint learning and joint inference can produce better prediction performance. We consider joint learning with separate inference, where only content features or discourse features are used for prediction (Separate-Inference). We further study learning separate classifiers for content selection and discourse relations without joint features (Separate-Learn).

We first show the phrase selection and discourse relation prediction results on \textsc{AMI-sub} in Tables~\ref{tab:ami_disc_phrase} and~\ref{tab:ami_disc_discourse}. As shown in Table~\ref{tab:ami_disc_phrase}, our models, trained with gold-standard discourse relations or latent ones with true attachment structure, yield significant better accuracy and F1 scores than SVM-based classifiers trained with the same feature sets for phrase selection (paired $t$-test, $p<0.05$). 
Our joint learning model with separate inference also outperforms neural network-based discourse parsing model~\cite{ji-haffari-eisenstein:2016:N16-1} in Table~\ref{tab:ami_disc_discourse}.

Moreover, Tables~\ref{tab:ami_disc_phrase} and~\ref{tab:ami_disc_discourse} demonstrate that joint learning usually produces superior performance for both tasks than separate learning. Combined with joint inference, our model obtains the best accuracy and F1 on phrase selection. 
This indicates that leveraging the interplay between content and discourse boost the prediction performance.  
Similar results are achieved on \textsc{AMI-full} and \textsc{ICSI-full} in Table~\ref{tab:ami_topic}, where latent discourse relations without true attachment structure are employed for training.

\begin{table}[t]
{
	\centering
	\fontsize{8}{9}\selectfont
    \setlength{\tabcolsep}{0.8mm}
	\begin{tabular}{|l|l|l|l|l|}
 	\hline
 		& \multicolumn{2}{c|}{\textsc{AMI-full}} & \multicolumn{2}{c|}{\textsc{ICSI-full}} \\
  		& \textbf{Acc} & \textbf{F1} & \textbf{Acc} & \textbf{F1} \\
 	\hline
 	\underline{\textbf{Comparisons}} & & & & \\
 
 	Baseline (Majority) & 61.8 & 38.2 & {\bf 75.3} & 43.0\\ 
 	SVM (with content features in \S~\ref{sec:feature}) &  58.6 & 56.7 &  66.2 & 53.1 \\ 
	\hline
	\hline 	
	
	\underline{\textbf{Our Models (Latent Discourse)}} & & & & \\
	Joint-Learn + Joint-Inference & {\bf 63.4}$\ast$ & {\bf 63.0}$\ast$ & 73.5$\ast$ & 61.4$\ast$ \\ 
	Joint-Learn + Separate-Inference & 57.7 & 57.5 & 70.0$\ast$ & {\bf 62.7}$\ast$ \\ 

%

 \hline
\end{tabular}

}
\caption{\fontsize{10}{12}\selectfont  Phrase-based content selection performance on \textsc{AMI-full} and \textsc{ICSI-full}. We display results of our models trained with latent discourse relations. Results that are significantly better than SVM-based model are highlighted with $\ast$ ($p < 0.05$, paired $t$-test). }
\label{tab:ami_topic}
\end{table}

\subsection{Phrase-Based Extractive Summarization}
We further evaluate whether the prediction of the content selection component can be used for summarizing the key points on discussion level. 
For each discussion, salient phrases identified by our model are concatenated in sequence for use as the summary. We consider two types of gold-standard summaries. One is utterance-level extractive summary, which consists of human labeled summary-worthy utterances. The other is abstractive summary, where we collect human abstract with at least one link from summary-worthy utterances.

\begin{table}[t]
\centering
	\fontsize{8}{9}\selectfont
    \setlength{\tabcolsep}{0.8mm}
\begin{tabular}{|l|l|l|l|l|l|l|l|}
\hline

	\multicolumn{8}{|l|}{\textit{Extractive Summaries as Gold-Standard}}\\ \hline
    & &\multicolumn{3}{c|}{\textsc{ROUGE-1}} & \multicolumn{3}{c|}{\textsc{ROUGE-SU4}}\\ \hline
    & Len & Prec & Rec & F1 & Prec & Rec & F1\\ 
	Longest DA & 30.9 & 64.4 & 15.0  & 23.1 & 58.6 & 9.3 & 15.3 \\ 
   	Centroid DA & 17.5 & {\bf 73.9} & 13.4 & 20.8 & {\bf 62.5} & 6.9 & 11.3 \\ 
	SVM & 49.8& 47.1 & 24.1 & 27.5 & 22.7 & 10.7 & 11.8 \\ 
    \newcite{liu2009unsupervised} & 62.4& 40.4 & 39.2 & 36.2 & 15.5 & 15.2 & 13.5 \\ 
    \hline\hline
	Our Model & 66.6 & 45.4 & 44.7 & 41.1$\ast$ & 24.1$\ast$ & 23.4$\ast$ & 20.9$\ast$ \\ 
   	Our Model-latent & 85.9 & 42.9 & {\bf 49.3} & {\bf 42.4}$\ast$ & 21.6 & {\bf 25.7}$\ast$ & {\bf 21.3}$\ast$ \\ 
   
 \hline\hline

	\multicolumn{8}{|l|}{\textit{Abstractive Summaries as Gold-Standard}}\\ \hline
    & &\multicolumn{3}{c|}{\textsc{ROUGE1}} & \multicolumn{3}{c|}{\textsc{ROUGE-SU4}}\\ \hline
    &  Len & Prec & Rec & F1 & Prec & Rec & F1\\ 
	Longest DA & 30.9 & 14.8  & 5.5 & 7.4 & 4.8 & 1.4 & 1.9 \\ 
   	Centroid DA & 17.5 &{\bf 24.9} &  5.6 & 8.5 & {\bf 11.6} & 1.4 & 2.2 \\ 
   	SVM & 49.8& 13.3& 9.7  & 9.5 & 4.4 & 2.4 & 2.4 \\ 
    \newcite{liu2009unsupervised} & 62.4& 10.3 & 16.7 & 11.3 & 2.7 & 4.5 & 2.8 \\ 
    \hline\hline
    
	Our Model & 66.6& 12.6 & 18.9 & {\bf 13.1}$\ast$ & 3.8 & 5.5$\ast$ & {\bf 3.7}$\ast$ \\ 
	Our Model-latent & 85.9 & 11.4 & {\bf 20.0} & 12.4$\ast$ & 3.3 & {\bf 6.1}$\ast$ & 3.5$\ast$ \\ 
   
    \hline
\end{tabular}
\caption{\fontsize{10}{12}\selectfont ROUGE scores for phrase-based extractive summarization evaluated against human-constructed utterance-level extractive summaries and abstractive summaries. Our models that statistically significantly outperform SVM and \newcite{liu2009unsupervised} are highlighted with $\ast$ ($p < 0.05$, paired $t$-test). Best ROUGE score for each column is in \textbf{bold}.}
\label{tab:summ}
\end{table}

\begin{figure}[t]
	{\fontsize{9}{10}\selectfont
    \setlength{\tabcolsep}{0.6mm}
    \begin{tabular}{|p{75mm}|}
    \hline
	
	\textbf{Meeting Clip}: \\
	D: can we uh power a light in this? can we get a strong enough battery to power a light? \\
	A: um i think we could because the lcd panel requires power, and the lcd is a form of a light so that$\ldots$\\
	D: $\ldots$it's gonna have to have something high-tech about it and that's gonna take battery power$\ldots$ \\
	D: illuminate the buttons. yeah it glows.\\
	D: well m i'm thinking along the lines of you're you're in the dark watching a dvd and you um you find the thing in the dark and you go like this $\ldots$ oh where's the volume button in the dark, and uh y you just touch it $\ldots$ and it lights up or something.\\

	\hline \hline    
    
	\textbf{Abstract by Human}: \\
    What sort of battery to use. The industrial designer presented options for materials, components, and batteries and discussed the restrictions involved in using certain materials.\\
	
	\hline \hline
	
	\textbf{Longest DA}: \\
	well m i'm thinking along the lines of you're you're in the dark watching a dvd and you um you find the thing in the dark and you go like this.\\
	\textbf{Centroid DA}: \\
	can we uh power a light in this?\\

	\textbf{Our Method}: \\
	- power a light, a strong enough battery, \\
	- requires power, a form, \\
	- a really good battery, battery power, \\
	- illuminate the buttons, glows, \\
	- watching a dvd, the volume button, lights up or something\\	
    \hline
    \end{tabular}
    
    }
	\caption{\fontsize{10}{12}\selectfont  Sample summaries output by different systems for a meeting clip from AMI corpus (less relevant utterances in between are removed). Salient phrases by our system output are displayed for each turn of the clip, with duplicated phrases removed for brevity. }
\label{fig:example_summary}
\end{figure}

We calculate scores based on ROUGE~\cite{Lin:2003:AES:1073445.1073465}, which is a popular tool for evaluating text summarization~\cite{gillick2009global,liu2010using}. ROUGE-1 (unigrams) and ROUGE-SU4 (skip-bigrams with at most 4 words in between) are used. 
Following previous work on meeting summarization~\cite{Riedhammer:2010:LSS:1837521.1837625,wang-cardie:2013:ACL2013}, we consider two dialogue act-level summarization baselines: (1) \textsc{longest DA} in each discussion is selected as the summary, and (2) \textsc{centroid DA}, the one with the highest TF-IDF similarity with all DAs in the discussion. 
We also compare with an unsupervised keyword extraction approach by \newcite{liu2009unsupervised}, where word importance is estimated by its TF-IDF score, POS tag, and the salience of its corresponding sentence. With the same candidate phrases as in our model, we extend \newcite{liu2009unsupervised} by scoring each phrase based on its average score of the words. Top phrases, with the same number of phrases output by our model, are included into the summaries. 
Finally, we compare with summaries consisting of salient phrases predicted by an SVM classifier trained with our content features.

From the results in Table~\ref{tab:summ}, we can see that phrase-based extractive summarization methods can yield better ROUGE scores for recall and F1 than baselines that extract the whole sentences. 
Meanwhile, our system significantly outperforms the SVM-based classifiers when evaluated on ROUGE recall and F1, while achieving comparable precision. Compared to \newcite{liu2009unsupervised}, our system also yields better results on all metrics.

Sample summaries by our model along with two baselines are displayed in Figure~\ref{fig:example_summary}. Utterance-level extract-based baselines unavoidably contain disfluency and unnecessary details. Our phrase-based extractive summary is able to capture the key points from both the argumentation process and important outcomes of the conversation. This implies that our model output can be used as input for an abstractive summarization system. It can also facilitate the visualization of decision-making processes.

\subsection{Further Analysis and Discussions}
\noindent \textbf{Features Analysis.} 
We first discuss salient features with top weights learned by our joint model. For content features, main speaker tends to utter more salient content. Higher TF-IDF scores also indicate important phrases. If a phrase is mentioned in previous turn and repeated in the current turn, it is likely to be a key point. 
For discourse features, structure features matter the most. For instance, jointly modeling the discourse relation of the parent node along with the current node can lead to better inference. An example is that giving more details on the proposal (\textsc{elaboration}) tends to lead to \textsc{positive} feedback. Moreover, \textsc{request} usually appears close to the root of the argument diagram tree, while \textsc{positive} feedback is usually observed on leaves. Adjacency pairs also play an important role for discourse prediction. 
For joint features, features that composite ``phrase mentioned in previous turn" and relation \textsc{positive} feedback or \textsc{request} yield higher weight, which are indicators for both key phrases and discourse relations. We also find that main speaker information composite with \textsc{elaboration} and \textsc{uncertain} are associated with high weights.

\noindent \textbf{Error Analysis and Potential Directions.} 
Taking a closer look at our prediction results, one major source of incorrect prediction for phrase selection is based on the fact that similar concepts might be expressed in different ways, and our model predicts inconsistently for different variations. For example, participants use both ``thick" and ``two centimeters" to talk about the desired shape of a remote control. However, our model does not group them into the same cluster and later makes different predictions. For future work, semantic similarity with context information can be leveraged to produce better clustering results. 
%
Furthermore, identifying discourse relations in dialogues is still a challenging task. For instance, 
``I wouldn't choose a plastic case" should be labeled as \textsc{option exclusion}, if the previous turns talk about different options. Otherwise, it can be labeled as \textsc{negative}. Therefore, models that better handle semantics and context need to be considered.

\section{Predicting Consistency of Understanding}
\label{sec:consistency}

As discussed in previous work~\cite{mulder2002assessing,mercer2004sociocultural}, both content and discourse structure are critical for building shared understanding among discussants. 
In this section, we test whether our joint model can be utilized to predict the consistency among team members' understanding of their group decisions, which is defined as consistency of understanding (COU) in \newcite{kim2016improving}. 

\newcite{kim2016improving} establish gold-standard COU labels on a portion of AMI discussions, by comparing participant summaries to determine whether participants report the same decisions. If all decision points are consistent, the associated topic discussion is labeled as \textit{consistent}; otherwise, the discussion is identified as \textit{inconsistent}. Their annotation covers the \textsc{AMI-sub} dataset. Therefore, we run the prediction experiments on \textsc{AMI-sub} by using the same annotation. 
Out of total 129 discussions in \textsc{AMI-sub}, 86 discussions are labeled as consistent and 43 are inconsistent.



We construct three types of features by using our model's predicted labels. Firstly, we learn two versions of our model based on the ``consistent" discussions and the ``inconsistent" ones in the training set, with learned parameters $\mathbf{w_{con}}$ and $\mathbf{w_{incon}}$. For a discussion in the test set, these two models output two probabilities $p_{con}=\max_{\mathbf{c}, \mathbf{d}} P(\mathbf{c}, \mathbf{d}|\mathbf{x}, \mathbf{w_{con}})$ and $p_{incon}=\max_{\mathbf{c}, \mathbf{d}} P(\mathbf{c}, \mathbf{d}|\mathbf{x}, \mathbf{w_{incon}})$. We use $p_{con}-p_{incon}$ as a feature. 

Furthermore, we consider discourse relations of length one and two from the discourse structure tree. Intuitively, some discourse relations, e.g., \textsc{elaboration} followed by multiple \textsc{positive} feedback, imply consistent understanding. 

The third feature is based on word entrainment, which has been shown to correlate with task success for groups~\cite{nenkova2008high}. Using the formula in~\newcite{nenkova2008high}, we compute the average word entrainment between the main speaker who utters the most words and all the other participants. The content words in the salient phrases predicted by our model is considered for entrainment computation.
 

\begin{table}[t]
\centering
\fontsize{9}{10}\selectfont
\setlength{\tabcolsep}{1.5mm}
\begin{tabular}{|l|c|c|}
    \hline
        & \textbf{Acc} & \textbf{F1} \\ \hline
        \underline{\textbf{Comparisons}} && \\
        Baseline (Majority) & 66.7 & 40.0  \\ 
        Ngrams (SVM) & 51.2 & 50.6  \\ 
		\newcite{kim2016improving} & 60.5 & 50.5 \\ \hline\hline
		        
        \underline{\textbf{Features from Our Model}}&&  \\ 
        Consistency Probability (Prob) & 52.7 & 52.1 \\ 
        Discourse Relation (Disc) & 63.6 & 57.1$\ast$  \\ 
        Word Entrainment (Ent) & 60.5$\ast$ & 57.1$\ast$\\ 
        Prob + Disc+ Ent & {\bf 68.2}$\ast$ & {\bf 63.1}$\ast$  \\ \hline\hline

        \underline{\textbf{Oracles}}&&  \\ 
        Discourse Relation & 69.8 & 62.7 \\ 
        Word Entrainment & 61.2 & 57.8 \\ \hline
\end{tabular}
\caption{\fontsize{10}{12}\selectfont Consistency of Understanding (COU) prediction results on \textsc{AMI-sub}. Results that statistically significantly outperform ngrams-based baseline and \newcite{kim2016improving} are highlighted with $\ast$ ($p < 0.05$, paired $t$-test). 
For reference, we also show the prediction performance based on gold-standard discourse relations and phrase selection labels.}
\label{tab:consistency}
\end{table}

\noindent \textbf{Results.} 
Leave-one-out is used for experiments. For training, our features are constructed from gold-standard phrase and discourse labels. Predicted labels by our model is used for constructing features during testing. SVM-based classifier is used for experimenting with different sets of features output by our model. 
A majority class baseline is constructed as well. We also consider an SVM classifier trained with ngram features (unigrams and bigrams). Finally, we compare with the state-of-the-art method in~\newcite{kim2016improving}, where discourse-relevant features and head gesture features are utilized in Hidden Markov Models to predict the consistency label.

The results are displayed in Table~\ref{tab:consistency}. All SVMs trained with our features surpass the ngrams-based baseline. Especially, the discourse features, word entrainment feature, and the combination of the three, all significantly outperform the state-of-the-art system by \newcite{kim2016improving}.\footnote{We also experiment with other popular classifiers, e.g. logistic regression or decision tree, and similar trend is respected.}

\section{Conclusion}
\label{sec:conclusion}
We presented a joint model for performing phrase-level content selection and discourse relation prediction in spoken meetings. Experimental results on AMI and ICSI meeting corpora showed that our model can outperform state-of-the-art methods for both tasks. Further evaluation on the task of predicting consistency-of-understanding in meetings demonstrated that classifiers trained with features constructed from our model output produced superior performance compared to the state-of-the-art model. This provides an evidence of our model being successfully applied in other prediction tasks in spoken meetings.

\section*{Acknowledgments}
This work was supported in part by National Science Foundation Grant IIS-1566382 and a GPU gift from Nvidia. We thank three anonymous reviewers for their valuable suggestions on various aspects of this work.

\bibliography{meeting,additional}

\begin{thebibliography}{}
\expandafter\ifx\csname natexlab\endcsname\relax\def\natexlab#1{#1}\fi

\bibitem[{Bokaei et~al.(2016)Bokaei, Sameti, and Liu}]{bokaei2016extractive}
Mohammad~Hadi Bokaei, Hossein Sameti, and Yang Liu. 2016.
\newblock {Extractive Summarization of Multi-party Meetings Through Discourse
  Segmentation}.
\newblock {\em Natural Language Engineering\/} 22(01):41--72.

\bibitem[{Bui et~al.(2009)Bui, Frampton, Dowding, and
  Peters}]{Bui:2009:EDM:1708376.1708410}
Trung~H. Bui, Matthew Frampton, John Dowding, and Stanley Peters. 2009.
\newblock {Extracting Decisions from Multi-party Dialogue Using Directed
  Graphical Models and Semantic Similarity}.
\newblock In {\em Proceedings of the SIGDIAL 2009 Conference: The 10th Annual
  Meeting of the Special Interest Group on Discourse and Dialogue\/}.
  Association for Computational Linguistics, Stroudsburg, PA, USA, SIGDIAL '09,
  pages 235--243.

\bibitem[{Carletta et~al.(2006)Carletta, Ashby, Bourban, Flynn, Guillemot,
  Hain, Kadlec, Karaiskos, Kraaij, Kronenthal, Lathoud, Lincoln, Lisowska,
  McCowan, Post, Reidsma, and Wellner}]{ami}
Jean Carletta, Simone Ashby, Sebastien Bourban, Mike Flynn, Mael Guillemot,
  Thomas Hain, Jaroslav Kadlec, Vasilis Karaiskos, Wessel Kraaij, Melissa
  Kronenthal, Guillaume Lathoud, Mike Lincoln, Agnes Lisowska, Iain McCowan,
  Wilfried Post, Dennis Reidsma, and Pierre Wellner. 2006.
\newblock {The AMI Meeting Corpus: A Pre-announcement}.
\newblock In {\em Proceedings of the Second International Conference on Machine
  Learning for Multimodal Interaction\/}. Springer-Verlag, Berlin, Heidelberg,
  MLMI'05, pages 28--39.

\bibitem[{Cohen et~al.(2004)Cohen, Carvalho, and Mitchell}]{Cohen04learningto}
William~W. Cohen, Vitor~R. Carvalho, and Tom~M. Mitchell. 2004.
\newblock {Learning to Classify Email into ``Speech Acts'' }.
\newblock In Dekang Lin and Dekai Wu, editors, {\em Proceedings of the 2004
  Conference on Empirical Methods in Natural Language Processing\/}.
  Association for Computational Linguistics, Barcelona, Spain, pages 309--316.

\bibitem[{Dielmann and Renals(2008)}]{dielmann2008recognition}
Alfred Dielmann and Steve Renals. 2008.
\newblock {Recognition of Dialogue Acts in Multiparty Meetings Using a
  Switching DBN}.
\newblock {\em IEEE transactions on audio, speech, and language processing\/}
  16(7):1303--1314.

\bibitem[{Fern{\'a}ndez et~al.(2008)Fern{\'a}ndez, Frampton, Dowding,
  Adukuzhiyil, Ehlen, and Peters}]{FernandezFDAEP08}
Raquel Fern{\'a}ndez, Matthew Frampton, John Dowding, Anish Adukuzhiyil,
  Patrick Ehlen, and Stanley Peters. 2008.
\newblock {Identifying Relevant Phrases to Summarize Decisions in Spoken
  Meetings}.
\newblock In {\em INTERSPEECH\/}. pages 78--81.

\bibitem[{Galley(2006)}]{Galley:2006:SCR:1610075.1610126}
Michel Galley. 2006.
\newblock {A Skip-chain Conditional Random Field for Ranking Meeting Utterances
  by Importance}.
\newblock In {\em Proceedings of the 2006 Conference on Empirical Methods in
  Natural Language Processing\/}. Association for Computational Linguistics,
  Stroudsburg, PA, USA, EMNLP '06, pages 364--372.

\bibitem[{Ghosh et~al.(2014)Ghosh, Muresan, Wacholder, Aakhus, and
  Mitsui}]{ghosh2014analyzing}
Debanjan Ghosh, Smaranda Muresan, Nina Wacholder, Mark Aakhus, and Matthew
  Mitsui. 2014.
\newblock {Analyzing Argumentative Discourse Units in Online Interactions}.
\newblock In {\em Proceedings of the First Workshop on Argumentation Mining\/}.
  pages 39--48.

\bibitem[{Gillick et~al.(2009)Gillick, Riedhammer, Favre, and
  Hakkani-Tur}]{gillick2009global}
Dan Gillick, Korbinian Riedhammer, Benoit Favre, and Dilek Hakkani-Tur. 2009.
\newblock {A Global Optimization Framework for Meeting Summarization}.
\newblock In {\em Acoustics, Speech and Signal Processing, 2009. ICASSP 2009.
  IEEE International Conference on\/}. IEEE, pages 4769--4772.

\bibitem[{Hakkani-Tur(2009)}]{hakkani2009towards}
Dilek Hakkani-Tur. 2009.
\newblock {Towards Automatic Argument Diagramming of Multiparity Meetings}.
\newblock In {\em Acoustics, Speech and Signal Processing, 2009. ICASSP 2009.
  IEEE International Conference on\/}. IEEE, pages 4753--4756.

\bibitem[{Hernault et~al.(2010)Hernault, Prendinger, duVerle, and
  Ishizuka}]{HernaultPdI10}
Hugo Hernault, Helmut Prendinger, David~A. duVerle, and Mitsuru Ishizuka. 2010.
\newblock {{HILDA:} {A} Discourse Parser Using Support Vector Machine
  Classification}.
\newblock {\em Dialogue {\&} Discourse\/} 1(3):1--33.

\bibitem[{Janin et~al.(2003)Janin, Baron, Edwards, Ellis, Gelbart, Morgan,
  Peskin, Pfau, Shriberg, Stolcke et~al.}]{icsi}
Adam Janin, Don Baron, Jane Edwards, Dan Ellis, David Gelbart, Nelson Morgan,
  Barbara Peskin, Thilo Pfau, Elizabeth Shriberg, Andreas Stolcke, et~al. 2003.
\newblock {The ICSI Meeting Corpus}.
\newblock In {\em Acoustics, Speech, and Signal Processing, 2003.
  Proceedings.(ICASSP'03). 2003 IEEE International Conference on\/}. IEEE,
  volume~1, pages I--I.

\bibitem[{Ji et~al.(2016)Ji, Haffari, and
  Eisenstein}]{ji-haffari-eisenstein:2016:N16-1}
Yangfeng Ji, Gholamreza Haffari, and Jacob Eisenstein. 2016.
\newblock {A Latent Variable Recurrent Neural Network for Discourse-Driven
  Language Models}.
\newblock In {\em Proceedings of the 2016 Conference of the North American
  Chapter of the Association for Computational Linguistics: Human Language
  Technologies\/}. Association for Computational Linguistics, San Diego,
  California, pages 332--342.

\bibitem[{Kalchbrenner and Blunsom(2013)}]{kalchbrenner2013recurrent}
Nal Kalchbrenner and Phil Blunsom. 2013.
\newblock {Recurrent Convolutional Neural Networks for Discourse
  Compositionality}.
\newblock In {\em Proceedings of the Workshop on Continuous Vector Space Models
  and their Compositionality\/}. Association for Computational Linguistics,
  Sofia, Bulgaria, pages 119--126.

\bibitem[{Kim and Shah(2016)}]{kim2016improving}
Joseph Kim and Julie~A Shah. 2016.
\newblock {Improving Team's Consistency of Understanding in Meetings}.
\newblock {\em IEEE Transactions on Human-Machine Systems\/} 46(5):625--637.

\bibitem[{Kirschner et~al.(2012)Kirschner, Buckingham-Shum, and
  Carr}]{kirschner2012visualizing}
Paul~A Kirschner, Simon~J Buckingham-Shum, and Chad~S Carr. 2012.
\newblock {\em {Visualizing Argumentation: Software Tools for Collaborative and
  Educational Sense-making}\/}.
\newblock Springer Science \& Business Media.

\bibitem[{Klein and Manning(2003)}]{Klein:2003:AUP:1075096.1075150}
Dan Klein and Christopher~D. Manning. 2003.
\newblock {Accurate Unlexicalized Parsing}.
\newblock In {\em Proceedings of the 41st Annual Meeting on Association for
  Computational Linguistics - Volume 1\/}. Association for Computational
  Linguistics, Stroudsburg, PA, USA, ACL '03, pages 423--430.

\bibitem[{Lin and Hovy(2003)}]{Lin:2003:AES:1073445.1073465}
Chin-Yew Lin and Eduard Hovy. 2003.
\newblock {Automatic Evaluation of Summaries Using N-gram Co-occurrence
  Statistics}.
\newblock In {\em Proceedings of the 2003 Conference of the North American
  Chapter of the Association for Computational Linguistics on Human Language
  Technology - Volume 1\/}. pages 71--78.

\bibitem[{Liu and Liu(2010)}]{liu2010using}
Fei Liu and Yang Liu. 2010.
\newblock {Using Spoken Utterance Compression for Meeting Summarization: A
  Pilot Study}.
\newblock In {\em Spoken Language Technology Workshop (SLT), 2010 IEEE\/}.
  IEEE, pages 37--42.

\bibitem[{Liu et~al.(2009)Liu, Pennell, Liu, and Liu}]{liu2009unsupervised}
Feifan Liu, Deana Pennell, Fei Liu, and Yang Liu. 2009.
\newblock {Unsupervised Approaches for Automatic Keyword Extraction Using
  Meeting Transcripts}.
\newblock In {\em Proceedings of Human Language Technologies: The 2009 Annual
  Conference of the North American Chapter of the Association for Computational
  Linguistics\/}. Association for Computational Linguistics, Boulder, Colorado,
  pages 620--628.

\bibitem[{Liu et~al.(2015)Liu, Shang, Wang, Ren, and Han}]{liu2015mining}
Jialu Liu, Jingbo Shang, Chi Wang, Xiang Ren, and Jiawei Han. 2015.
\newblock Mining quality phrases from massive text corpora.
\newblock In {\em Proceedings of the 2015 ACM SIGMOD International Conference
  on Management of Data\/}. ACM, pages 1729--1744.

\bibitem[{Loza et~al.(2014)Loza, Lahiri, Mihalcea, and Lai}]{loza2014building}
Vanessa Loza, Shibamouli Lahiri, Rada Mihalcea, and Po-Hsiang Lai. 2014.
\newblock {Building a Dataset for Summarization and Keyword Extraction from
  Emails}.
\newblock In {\em LREC\/}. pages 2441--2446.

\bibitem[{McKeown et~al.(2007)McKeown, Shrestha, and Rambow}]{mckeown2007using}
Kathleen McKeown, Lokesh Shrestha, and Owen Rambow. 2007.
\newblock {Using Question-answer Pairs in Extractive Summarization of Email
  Conversations}.
\newblock In {\em International Conference on Intelligent Text Processing and
  Computational Linguistics\/}. Springer, pages 542--550.

\bibitem[{Mehdad et~al.(2014)Mehdad, Carenini, and
  Ng}]{mehdad-carenini-ng:2014:P14-1}
Yashar Mehdad, Giuseppe Carenini, and Raymond~T. Ng. 2014.
\newblock {Abstractive Summarization of Spoken and Written Conversations Based
  on Phrasal Queries}.
\newblock In {\em Proceedings of the 52nd Annual Meeting of the Association for
  Computational Linguistics (Volume 1: Long Papers)\/}. Association for
  Computational Linguistics, Baltimore, Maryland, pages 1220--1230.

\bibitem[{Mercer(2004)}]{mercer2004sociocultural}
Neil Mercer. 2004.
\newblock {Sociocultural Discourse Analysis}.
\newblock {\em Journal of applied linguistics\/} 1(2):137--168.

\bibitem[{Mulder et~al.(2002)Mulder, Swaak, and Kessels}]{mulder2002assessing}
Ingrid Mulder, Janine Swaak, and Joseph Kessels. 2002.
\newblock {Assessing Group Learning and Shared Understanding in
  Technology-mediated Interaction}.
\newblock {\em Educational Technology \& Society\/} 5(1):35--47.

\bibitem[{Murray et~al.(2010)Murray, Carenini, and
  Ng}]{Murray:2010:GVA:1873738.1873753}
Gabriel Murray, Giuseppe Carenini, and Raymond Ng. 2010.
\newblock {Generating and Validating Abstracts of Meeting Conversations: A User
  Study}.
\newblock In {\em Proceedings of the 6th International Natural Language
  Generation Conference\/}. Association for Computational Linguistics,
  Stroudsburg, PA, USA, INLG '10, pages 105--113.

\bibitem[{Murray et~al.(2006)Murray, Renals, Carletta, and
  Moore}]{murray2006incorporating}
Gabriel Murray, Steve Renals, Jean Carletta, and Johanna Moore. 2006.
\newblock {Incorporating Speaker and Discourse Features into Speech
  Summarization}.
\newblock In {\em Proceedings of the main conference on Human Language
  Technology Conference of the North American Chapter of the Association of
  Computational Linguistics\/}. Association for Computational Linguistics,
  pages 367--374.

\bibitem[{Nenkova et~al.(2008)Nenkova, Gravano, and
  Hirschberg}]{nenkova2008high}
Ani Nenkova, Agustin Gravano, and Julia Hirschberg. 2008.
\newblock {High Frequency Word Entrainment in Spoken Dialogue}.
\newblock In {\em Proceedings of the 46th annual meeting of the association for
  computational linguistics on human language technologies: Short papers\/}.
  Association for Computational Linguistics, pages 169--172.

\bibitem[{Oya and Carenini(2014)}]{oya2014extractive}
Tatsuro Oya and Giuseppe Carenini. 2014.
\newblock {Extractive Summarization and Dialogue Act Modeling on Email Threads:
  An Integrated Probabilistic Approach}.
\newblock In {\em 15th Annual Meeting of the Special Interest Group on
  Discourse and Dialogue\/}. page 133.

\bibitem[{Perret et~al.(2016)Perret, Afantenos, Asher, and
  Morey}]{perret-EtAl:2016:N16-1}
J\'{e}r\'{e}my Perret, Stergos Afantenos, Nicholas Asher, and Mathieu Morey.
  2016.
\newblock {Integer Linear Programming for Discourse Parsing}.
\newblock In {\em Proceedings of the 2016 Conference of the North American
  Chapter of the Association for Computational Linguistics: Human Language
  Technologies\/}. Association for Computational Linguistics, San Diego,
  California, pages 99--109.

\bibitem[{Riedhammer et~al.(2010)Riedhammer, Favre, and
  Hakkani-T\"{u}r}]{Riedhammer:2010:LSS:1837521.1837625}
Korbinian Riedhammer, Benoit Favre, and Dilek Hakkani-T\"{u}r. 2010.
\newblock {Long Story Short - Global Unsupervised Models for Keyphrase Based
  Meeting Summarization}.
\newblock {\em Speech Commun.\/} 52(10):801--815.

\bibitem[{{Rienks} et~al.(2005){Rienks}, {Heylen}, and van~der
  {Weijden}}]{so65562}
Rutger {Rienks}, Dirk {Heylen}, and E.~van~der {Weijden}. 2005.
\newblock {Argument Diagramming of Meeting Conversations}.
\newblock In A.~{Vinciarelli} and J-M. {Odobez}, editors, {\em International
  Workshop on Multimodal Multiparty Meeting Processing, MMMP 2005, part of the
  7th International Conference on Multimodal Interfaces, ICMI 2005\/}.

\bibitem[{Rohanimanesh et~al.(2011)Rohanimanesh, Bellare, Culotta, McCallum,
  and Wick}]{rohanimanesh2011samplerank}
Khashayar Rohanimanesh, Kedar Bellare, Aron Culotta, Andrew McCallum, and
  Michael~L Wick. 2011.
\newblock {Samplerank: Training Factor Graphs with Atomic Gradients}.
\newblock In {\em Proceedings of the 28th International Conference on Machine
  Learning (ICML-11)\/}. pages 777--784.

\bibitem[{Smith and Eisner(2008)}]{smith2008dependency}
David~A Smith and Jason Eisner. 2008.
\newblock {Dependency Parsing by Belief Propagation}.
\newblock In {\em Proceedings of the Conference on Empirical Methods in Natural
  Language Processing\/}. Association for Computational Linguistics, pages
  145--156.

\bibitem[{Stolcke et~al.(2000)Stolcke, Ries, Coccaro, Shriberg, Bates,
  Jurafsky, Taylor, Martin, Van Ess-Dykema, and Meteer}]{stolcke2000dialogue}
Andreas Stolcke, Klaus Ries, Noah Coccaro, Elizabeth Shriberg, Rebecca Bates,
  Daniel Jurafsky, Paul Taylor, Rachel Martin, Carol Van Ess-Dykema, and Marie
  Meteer. 2000.
\newblock {Dialogue Act Modeling for Automatic Tagging and Recognition of
  Conversational Speech}.
\newblock {\em Computational linguistics\/} 26(3):339--373.

\bibitem[{Wang and Cardie(2012)}]{wang-cardie:2012:SIGDIAL20122}
Lu~Wang and Claire Cardie. 2012.
\newblock {Focused Meeting Summarization via Unsupervised Relation Extraction}.
\newblock In {\em Proceedings of the 13th Annual Meeting of the Special
  Interest Group on Discourse and Dialogue\/}. Association for Computational
  Linguistics, Seoul, South Korea.

\bibitem[{Wang and Cardie(2013)}]{wang-cardie:2013:ACL2013}
Lu~Wang and Claire Cardie. 2013.
\newblock {Domain-Independent Abstract Generation for Focused Meeting
  Summarization}.
\newblock In {\em Proceedings of the 51st Annual Meeting of the Association for
  Computational Linguistics (Volume 1: Long Papers)\/}. Association for
  Computational Linguistics, Sofia, Bulgaria, pages 1395--1405.

\end{thebibliography}
\bibliographystyle{acl_natbib}

\appendix

\end{document}